\documentclass[11pt]{article}
\usepackage{coling2018}
\usepackage{times}
\usepackage{url}
\usepackage{latexsym}
\usepackage{amsmath}
\usepackage{amsfonts}
\usepackage{algorithm}
\usepackage{algorithmic}

\usepackage{booktabs}
\usepackage{tikz}
\usepackage{lipsum}
\usepackage{ntheorem}
\usepackage{makecell}
\usepackage{multirow}
\usepackage{array}
\usepackage{bbm}
\usepackage{multicol}
\usepackage{blindtext}
\usepackage[labelfont={small,bf},font={small,bf},skip=0pt]{caption}
\usepackage{caption}
\usepackage{subcaption}

\DeclareMathOperator*{\argmax}{arg\,max}

\title{A New Concept of Deep Reinforcement Learning based Augmented General Sequence Tagging System}
 
\author{Yu Wang$^\dag$ \\
  Samsung Research America
  \\\And
  Abhishek Patel$^\ddag$ \\
  Samsung Research America\\
   \tt\small{$^\dag$:$\lbrace$yu.wang1, hongxia.jin$\rbrace$@samsung.com,$^\ddag$:abhishek.p@partner.samsung.com} \\\And
   Hongxia Jin$^\dag$\\
  Samsung Research America\\}
\date{}

\begin{document}
\maketitle
\begin{abstract}
In this paper, a new deep reinforcement learning based augmented general sequence tagging system is proposed. The new system contains two parts: a deep neural network (DNN) based sequence tagging model and a deep reinforcement learning (DRL) based augmented tagger. The augmented tagger helps improve system performance by modeling the data with minority tags. The new system is evaluated on SLU and NLU sequence tagging tasks using ATIS and CoNLL-2003 benchmark datasets, to demonstrate the new system's outstanding performance on general tagging tasks. Evaluated by F1 scores, it shows that the new system outperforms the current state-of-the-art model on ATIS dataset by 1.9 $\%$ and that on CoNLL-2003 dataset by 1.4 $\%$.
\end{abstract}

\section{Introduction}
\blfootnote{
    %
    %
    \hspace{-0.65cm}  
This work is licensed under a Creative Commons Attribution 4.0 International License. License details: http://creativecommons.org/licenses/by/4.0/

    %
    %
    %
    %
}
Sequence tagging/labeling is one of the general techniques required for implementing different natural/spoken language understanding (NLU/SLU) tasks, such as: Named-entity recognition (NER), Part-of-speech (POS) Tagging in NLU, or Slot filling in SLU. Though the purpose of the tasks and their sequence labels are different, most of the state-of-the-art results for these tasks are generated using neural network based architectures. For example, Ma et al. demonstrates that their BLSTM-CNN-CRF model \cite{ma2016end} can achieve state-of-the-art performance for the NER task (without using extra features) on CoNLL-2003 dataset; Liu et al. gives an attention-based bi-directional LSTM model \cite{liu2016attention} for the slot-filling task, which also gives the best performance on the popular ATIS dataset \cite{hemphill1990atis} for SLU tasks (This result is obtained by intial paper submission, a better result is published later by Yu et al. \cite{wang2018bi}). The details of these related works will be given in next background section. The solution for general sequence tagging tasks (like NER/Slot filling) should mainly contain two properties: the first one is that the model can handle the language sequence in a word/token level manner, \emph{i.e.}, it can read in the sequence word by word and outputs their corresponding labels; the second property is that the model should capture the contextual information/constraints which helps understand the word/token's tags, either as named-entity or simply slot tags. In\cite{lample2016neural,ma2016end}, the authors use a conditional random filed (CRF) to assist their basic bidirectional LSTM model to capture the contextual constraints in NER task. Similarly, Liu et al. uses an attention based mechanism to take advantages of the weighted collective information from the hidden states of all words in an utterance. 

Due to the imbalance of data distribution under different labels, normally the evaluations for different models will be based on their precision and recall, or simply the F1 scores. It can be observed that, though the neural network based models are becoming more and more complicated, it becomes harder to make further improvement on these tagging tasks by using more advanced network structures. The main reasons are from two aspects:\\
1. Though the tagging models are designed to be more and more complicated, they also becomes more likely to be over-fitting by using more layers, hidden nodes or advance recurrent neural network structures (like bidirectional LSTM). A model's performance may easily reach a bottom-neck by simply tuning hyper-parameters or changing network structures.\\
2. Considering the fact that the numbers of words/tokens under some wrongly labeled tags may be quite small, it may not give us more improvement if we still use the same group of training data by simply changing the model. Normally a model will focus on finding the pattern of the data under majority tags instead of minority. Sometimes, even a new model can improve the performance on minority tags' classes, at the cost of degrading the model's performance on the majority tags with more training data, which is a common issue resulting from the imbalanced data distribution under different tags \cite{he2009learning,chawla2004special}.

Weighted sampling is one general solution to resolve the imbalanced data issue by giving more weights to minority tags and fewer to the majority ones \cite{he2009learning,sun2009classification}. The technique, however, gives the issue of distorting the original tagging distribution, which may adversely affect model's performance on majority tags. Similar approach, like adaptive synthetic sampling \cite{he2008adasyn},  also suffers from the same problem.

This gives us the dilemma on dealing with imbalanced tagged data (either for slot filling or NER task): On the one hand, it is a non-trivial task to re-organizing/sampling the data under different tags without any distortion of information; on the other hand, the improvement by changing models has an upper bound which is mainly decided by the data pattern itself instead of the model structure. Due to these two reasons, in this paper, a new concept of augmented tagging system based on deep reinforcement learning is proposed to furture improve the performances of sequence labeling tasks without sacrificing their original data distributions. This novel system will use a deep reinforcement learning based compensatory model to capture the wrong labeled tags and learn their correct labels, hence it can improve the whole model system performance without affecting the original correctly labeled tags. Detailed experiments will be further conducted on two sequence labeling tasks in the domain of SLU/NLU, as NER and slot filling, using public datasets.

The paper is organized as following: In section 2, a brief overview of two common NLP/SLU tagging tasks, \emph{i.e.} NER and Slot filling, are given. Their current state-of-the-art models will also be explained briefly. Section 3 gives a background of deep reinforcement learning (DRL) first, then illustrates our DRL based augmented general tagging system (DAT) in detail. In section 4, two different tagging tasks on different datasets are given, One is a slot filling task performed on the ATIS benchmark dataset. The other is a named-entity recognition (NER) task, tested on the CoNLL-2003 benchmark dataset. Both experiments will compare our new algorithm with their current best-performed models with state-of-the-art results.

\section{Background}
In this section, a brief background overview on the models for two sequence labeling tasks will be given. Their state-of-the-art neural network structures will also be discussed for further comparison purpose.

\subsection{Slot filling in SLU}
\label{sec-SLU}
\begin{figure}
\centering
\begin{minipage}{.49\textwidth}
  \centering
  \includegraphics[width=1\linewidth]{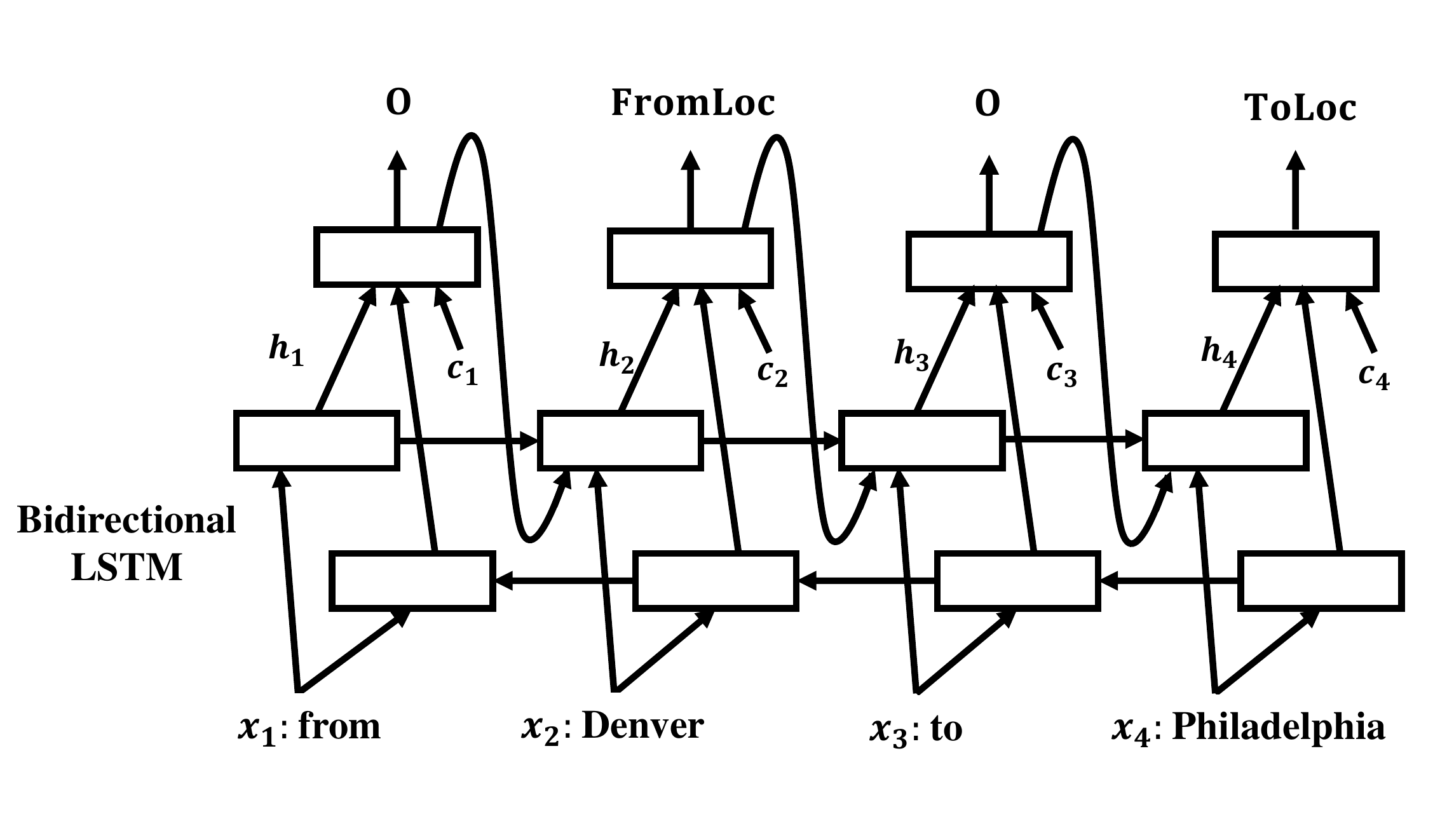}
  \caption{Attention based Bidirectional LSTM for Slot Filling}
  \label{fig:SlotTagger}
\end{minipage}%
\hspace{0.15cm}
\begin{minipage}{.49\textwidth}
  \centering
  \includegraphics[width=0.95\linewidth]{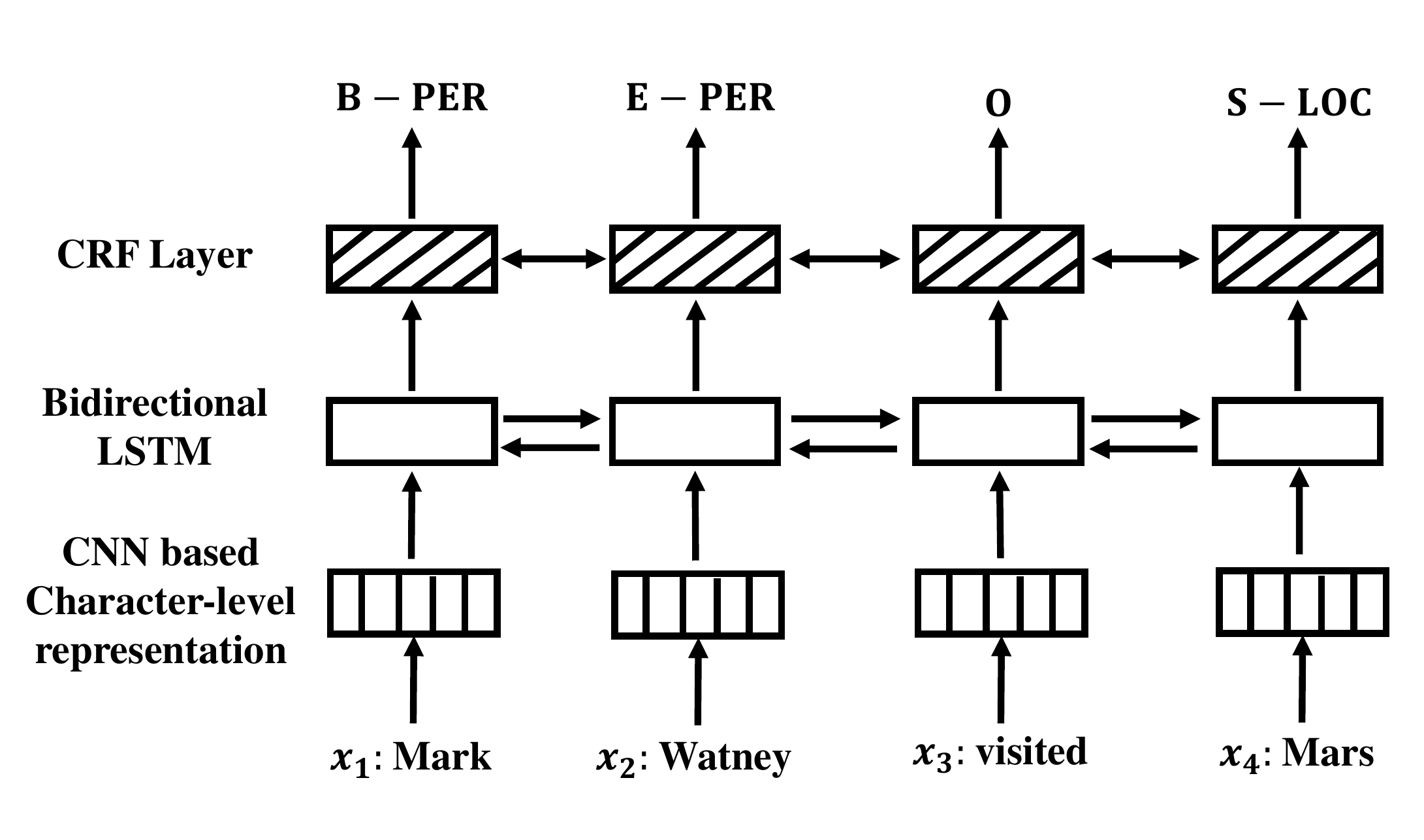}
  \caption{BLSTM-CNN-CRF Structure for Named-entity Recognition}
  \label{fig:NERTagger}
\end{minipage}
\end{figure}

Slot filling task is one of the fundamental tasks in spoken language understanding. It sequentially labels the words in an utterance using the pre-defined types of attributes or slot tags. The most straight-forward approach for this task is by using a single recurrent neural network (RNN) to generate sequential tags by reading an input sentence word by word. The current state-of-the-art result on ATIS is an attention based bidirectional LSTM model \cite{liu2016attention}. A brief overview about this approach is given for a better illustration of our new algorithm. The model introduced in \cite{liu2016attention} covers both of the intent detection and slot filling tasks. Considering the focus of this paper, we will only discuss the slot filling part of the model. The structure of the attention based BLSTM model for slot filling task is as shown in Figure \ref{fig:SlotTagger}. 

As shown in the figure, a contextual vector $c_i(\cdot)$ is defined using the attention of hidden states $h_j$:
\begin{equation}
c_i=\sum_{j=1}^L\alpha_{i,j}h_j
\end{equation}
where $\alpha_{i,j}$ is the attention coefficient defined as:
\begin{equation}
\begin{split}
\alpha_{i,j}&=\dfrac{e^{e_{i,j}}}{\sum_{k=1}^L e^{e_{i,k}}}\\
e_{i,k}&=\phi(s_{i-1},h_k)
\end{split}
\end{equation}
where $\phi(\cdot)$ is a feed-forward neural network and $s_{i-1}$ is the decoder hidden state.\\
This model consists of tw0 main properties: \\
1. It uses the Bi-direction LSTM (BLSTM) structure to capture the long-term dependencies in both directions.\\
2. The attention vector $c_i$ gives additional contextual information, for which cannot be captured by the hidden states in BLSTM.\\

The model gives the state-of-the-art slot filling performance on ATIS dataset. In this paper, we will also use this model as a baseline for comparison purpose. Also, this attention based model is chosen as the DNN part of our new system, \emph{i.e.}$f_{dnn}$, to generate the filtered input data to the deep reinforcement learning based augmented tagger (DAT) sub-model of the system.

\subsection{Named-entity recognition in NLU}
\label{sec-NER}
Another very common NLU sequence labeling task is named-entity recognition (NER), which seeks to locate and classify entities into some pre-defined named labels. In this section, a brief overview about the neural network based NER model as in \cite{ma2016end} will be given. The model gives the state-of-the-art performance on CoNLL-2003 dataset (without lexicon features). 

The structure contains a convolutional neural network for generating character-level representations, which is used as an input to a bidirectional LSTM network followed by a conditional random field layer. The conditional random field layer is designed to capture the strong dependency across the labels, which can be formulated as:
\begin{equation}
\begin{split}
s(X,y)&=\sum_{i=0}^nA_{y_i,y_{i+1}}+\sum_{i=0}^nP_{i,y_i}\\
p(y|X)&=\dfrac{e^{s(X,y)}}{\sum_{\tilde{y}\in Y_x}e^{s(X,\tilde{y})}}
\end{split}
\end{equation}
where $s(X,y)$ is the score defined a predicted labeling sequence $y$ given $X$. $A_{y_i,y_j}$ represents transition score from the tag $y_i$ to tag $y_j$, and $P_{i,y_i}$ is the probability of the $y_i$ output tag of the $i^{th}$ input word generated by the bidirectional LSTM. $Y_X$ represents all possible tag sequences for an input sentence $X$, and $p(y|X)$ is the predicted output $y$'s tagging probability giving an input words sequence $X=\lbrace x_1, x_2, \cdots, x_n \rbrace$.

The model shown above gives the state-of-the-art result on the English NER task with an F1 score of 91.2 on CoNLL-2003 dataset (without extra features).

Though decent experimental results are obtained by using deep learning models on the slot filling and NER tagging tasks, the potentials of these models are still limited by the imbalanced data distribution, since most of the tokens don't have any entity names (labeled as 'O'). The model can easily cover the majority tags but not the minority ones. Table \ref{table:data_comparison2} gives a summary of minority tags (each tag has less than 1$\%$ of entire data) and majority tags on ATIS and CoNLL-2003 datasets, and their corresponding occurrence ratio among the wrongly labeled data in test dataset.
\begin{table*}[t]\scriptsize
\centering
	\caption{Comparison between Minority Tags and Majority Tags on ATIS and CoNLL-2003}
	\label{table:data_comparison2}
	\begin{tabular}{>{\centering\arraybackslash}p{3cm}|>{\centering\arraybackslash}p{1cm}>{\centering\arraybackslash}p{1cm}|>{\centering\arraybackslash}p{1cm}>{\centering\arraybackslash}p{1cm}|>{\centering\arraybackslash}p{1cm}>{\centering\arraybackslash}p{1cm}|>{\centering\arraybackslash}p{1cm}>{\centering\arraybackslash}p{1cm}}
		\toprule
		
		\multicolumn{1}{c}{} & \multicolumn{4}{c|}{ATIS} & \multicolumn{4}{c}{\makecell{CoNLL-2003}}\\
		
	\multicolumn{1}{c}{} & \multicolumn{2}{c|}{\makecell{\textbf{Training Data}}} & \multicolumn{2}{c}{\makecell{\textbf{Test Data}}}&\multicolumn{2}{c|}{\makecell{\textbf{Training Data}}}&\multicolumn{2}{c}{\makecell{\textbf{Test Data}}}\\
		\midrule
		\multirow{2}{*} {}  & $\#$ of tag types&$\#$ of tokens&\multicolumn{2}{c|}{\makecell{$\%$ of wrongly labeled data}}& $\#$ of tag types&$\#$ of tokens& \multicolumn{2}{c}{\makecell{$\%$ of wrongly labeled data}}\\
		\midrule
		\multirow{2}{*} {}Minority Tags  & 119& 6,323&\multicolumn{2}{c|}{\makecell{92$\%$}}&2&2,312&\multicolumn{2}{c}{\makecell{78$\%$}}\\
		($<$1$\%$ of total $\#$ of data)\\
		\multirow{2}{*}{}Majority Tags  & 8 & 38,707&\multicolumn{2}{c|}{\makecell{8$\%$}}&7&202,255&\multicolumn{2}{c}{\makecell{22$\%$}} \\
		($\geq$ 1$\%$ of total $\#$ of data)\\

		\midrule
		\multirow{2}{*}{}Total & 127 &45,030&\multicolumn{2}{c|}{\makecell{100$\%$}}&9&204,567&\multicolumn{2}{c}{\makecell{100$\%$}}\\
		
		\bottomrule		
	\end{tabular}
\end{table*}

In Table \ref{table:data_comparison2}, it shows that about 92$\%$ of the wrongly labeled data are from minority tags in ATIS dataset and 78$\%$ in CoNLL03 dataset. This gives us an indication that, in order to further improve the model performance, the new model needs to figure out how to better label the minority tags and wrongly labeled tags without sacrificing the model performance on majority tags.

\section{Deep Reinforcement Learning}
In this section, a novel deep reinforcement learning (DRL) based augmented tagging system is proposed to address the issue as described earlier. The system contains two parts: one is the original deep neural network(NN) based tagger ($f_{dnn}$) as in section \ref{sec-SLU} for slot filling and section \ref{sec-NER} for NER, the second part is a deep reinforcement learning (DRL) based augmented tagger ($f_{dat}$), which can be trained to learn the correct tags that are labeled wrongly by the deep learning based tagger, such that it can compensate the weakness of the original single tagger. 

\subsection{Reinforcement Learning and Deep Q Network (DQN)}
Reinforcement learning (RL) is a bit different from the supervised learning and unsupervised learning algorithm. Formulated as a markov decision process (MDP) , an RL based model mainly contains several key elements: state ($s_t$), action ($a_t$), rewards ($r_t$) and policy ($\pi$). In a given a state of a stochastic environment, an agent is seeking its best action to perform in order to maximize its expected rewards to be obtained, by following some policy. The main target of a RL based model is to seek the best policy, hence the corresponding action, for an agent to perform. There are mainly three types of reinforcement learning algorithms: value-based, policy gradient and actor-critic. In our scenario, due to the necessity to define rewards at each state and small discrete action space, a value-based model design using $Q$-learning  is proposed for our problem. Its optimal action-value function $Q^\pi$, which is also the maximum expected reward obtained by selecting the best policy $\pi$, is defined as:
\begin{equation}
Q^{\pi}(s_t,a_t)=\max_{\pi} \mathbb{E}[R_t \mid a_t,s_t,\pi]
\end{equation} 

One well-known issue of applying reinforcement learning in real problems is that, in order to compute this optimal action-value function $Q^\pi$, it is necessary to store all of the $Q$ values for each state action pair ($s_t,a_t$) in a table, which is not very practical if the state or action space is large. This problem is described as ``the curse of dimensionality" by Bellman \cite{powell2007approximate}. One approach to overcome it is by training a deep neural network based estimator $Q(s_t,a_t|\theta_t)$ to estimate $Q^\pi(s_t,a_t)$ at time $t$. However, these types of estimators tend to be unstable since the convergence cannot be fully guaranteed. Recently, the deep $Q$ network (DQN) \cite{mnih2013playing,mnih2015human} demonstrates much better convergence performance on large state space, by taking advantage of a novel training technique called ``experience replay''. In this paper, we will use DQN as our selected deep reinforcement learning structure to satisfy the model requirement due to large state space.

In DQN structure, the optimal action-value function $Q^\pi(s_t,a_t)$ is estimated by a neural network based estimator $Q(s_t,a_t|\theta_t)$, \emph{i.e.}
\begin{equation}
Q(s_t,a_t|\theta_t)\sim Q^\pi(s_t,a_t)
\end{equation}
where $\theta_t$ is the neural network parameter, and the state $s_t$ serves as the input of the network. In next section we will describe how to use the techniques in DQN to build our augmented tagger from a system perspective. Also, the states, actions and rewards of the DRL based tagger are also going to be carefully designed next.

\subsection{DRL based Augmented Tagger (DAT) System}
\begin{figure}
\centering
\begin{minipage}{.49\textwidth}
  \centering
  \includegraphics[width=1\linewidth]{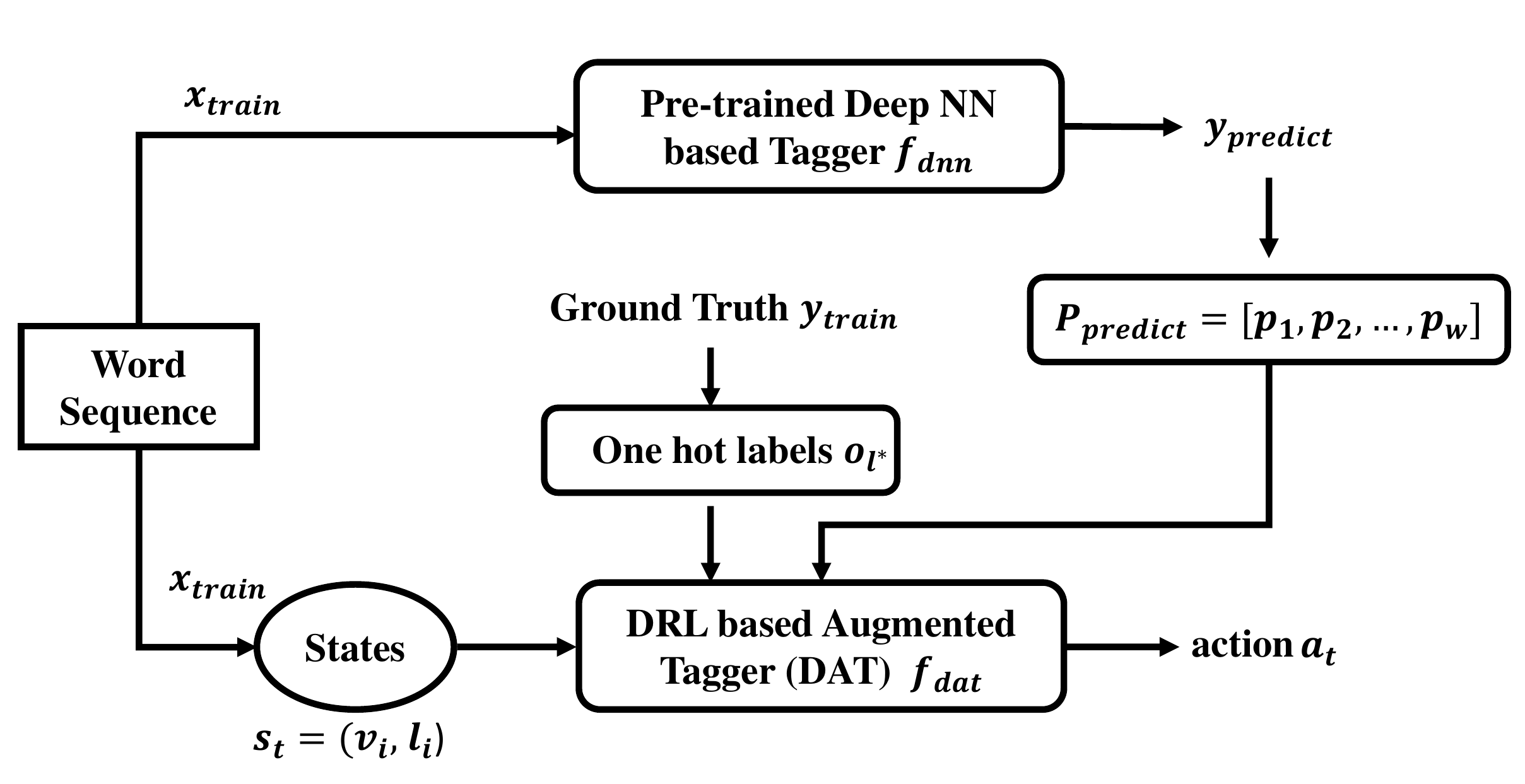}
  \caption{Training Model of DRL based Augmented Tagger}
  \label{fig:DATTraining}
\end{minipage}%
\hspace{0.15cm}
\begin{minipage}{.49\textwidth}
  \centering
  \includegraphics[width=0.95\linewidth]{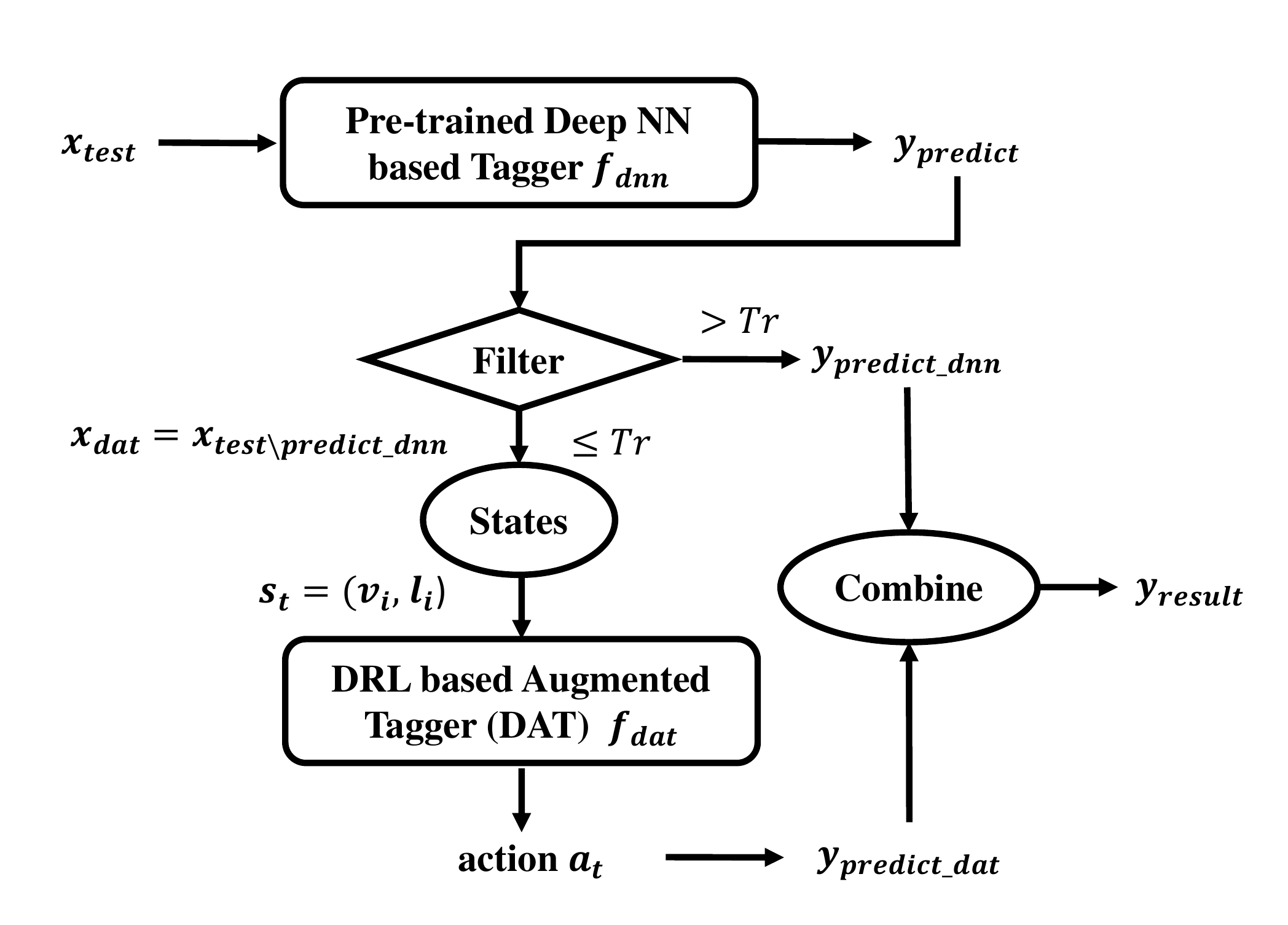}
  \caption{Inference Model of DRL based Augmented Tagger}
  \label{fig:DATInference}
\end{minipage}
\end{figure}

The design of DRL based augmented tagging (DAT) system is as given in Figure \ref{fig:DATTraining} and Figure \ref{fig:DATInference}. Due to the differences of using augmented tagger in training and inference, we discuss these two parts separately. 

\subsubsection{Training DAT System}
\begin{figure}
\centering
\begin{minipage}{.48\textwidth}
  \centering
  \includegraphics[width=1\linewidth]{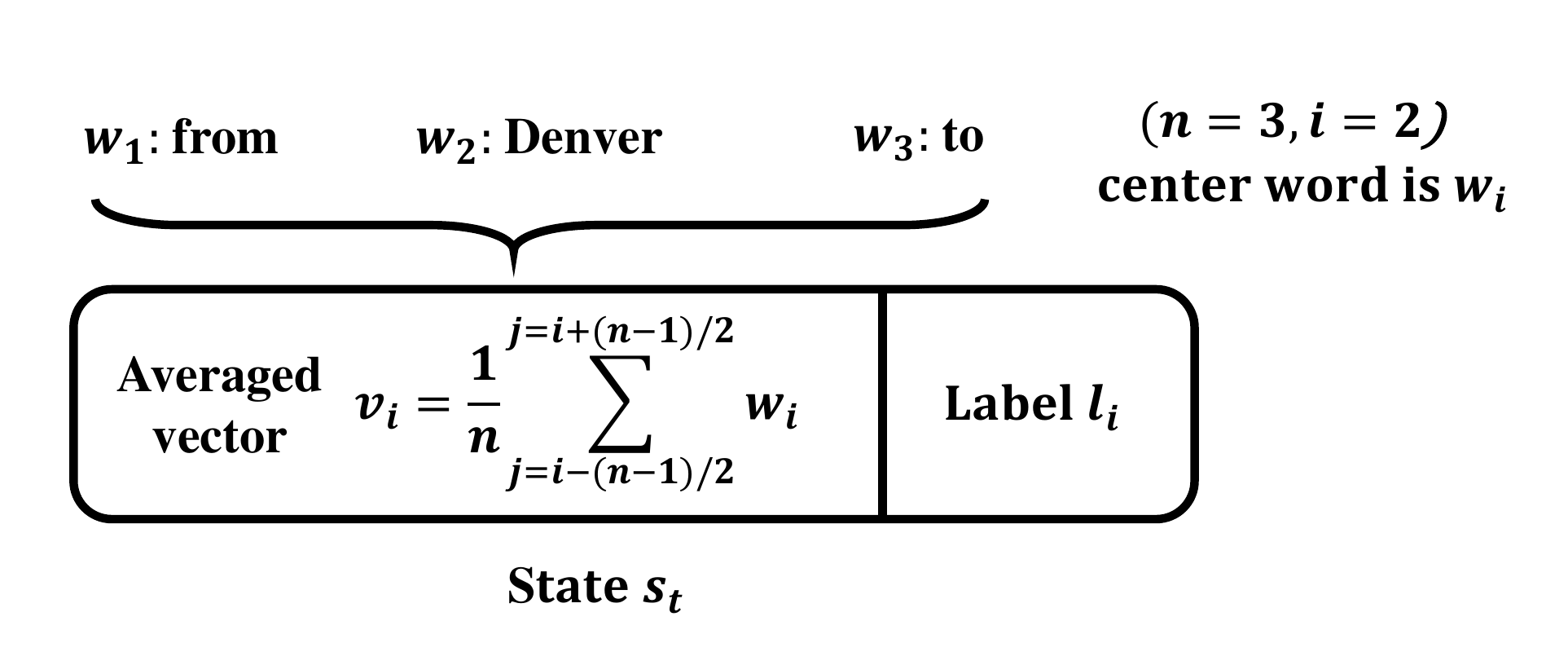}
  \caption{State of DAT}
  \label{fig:State}
\end{minipage}%
\hspace{0.15cm}
\begin{minipage}{.48\textwidth}
  \centering
  \includegraphics[width=0.95\linewidth]{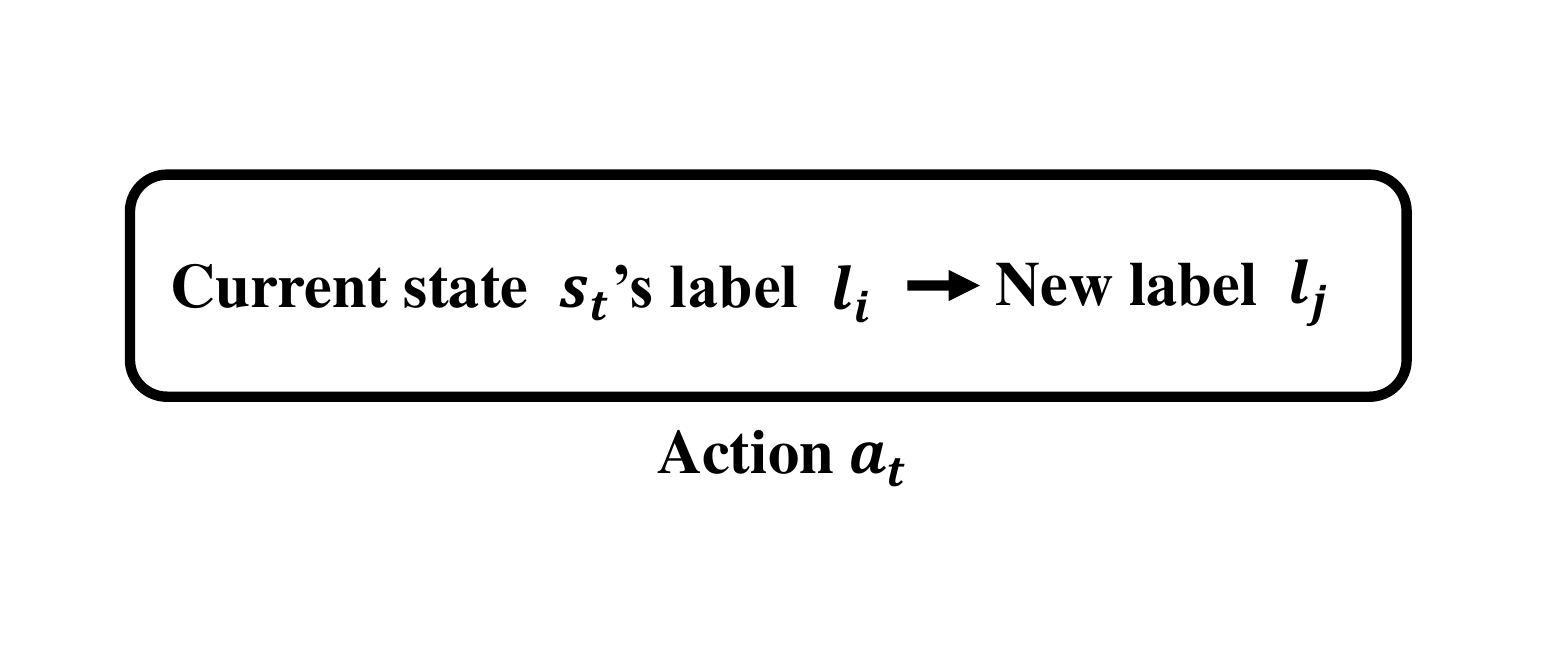}
  \caption{Action of DAT}
  \label{fig:Action}
\end{minipage}
\end{figure}

In order to train the entire tagging system, one may need to first pre-train a deep NN based tagger $f_{dnn}$, this tagger will follow the design as discussed in section \ref{sec-SLU} for slot filling task or section \ref{sec-NER} for the NER task. Once $f_{dnn}$ is fully trained, the output labels of $f_{dnn}$ using training input $x_{train}$ are stored as $y_{predict}$ with a probability distribution $P_{y_{predict}}=[p_1, p_2,\cdots, p_w]$, where $w$ is the total number of tags in the entire training dataset. Following is the DRL  based design of our new DAT structure:

{\bf{States ($s_t$):}} The DAT model's state $s_t$ is shown in Figure \ref{fig:State}. The state is defined by each word/token $w_i$ in sentences, it mainly contains two parts: the first part contains the word level information represented by an $n$-gram ($n$ is odd) averaged vector $v_i$, and the other part is a given label $l_i$ of $w_i$. During the generation of training states, $l_i$ uses all possible tags for the word/token $w_i$. $v_i$ is defined as the average of the vector of word sequences from $w_{i-(n-1)/2}$ to $w_{i+(n-1)/2}$, where $w_j$ is the center of this sequence:
\begin{equation}
v_i=\dfrac{1}{n}\sum_{j=i-(n-1)/2}^{j=i+(n-1)/2} w_j
\end{equation}
{\it{Remarks:} The reason to use an n-gram vector $v_i$ to substitute $w_i$ as a word level information in a state is due to that we want to better capture the contextual information compared to that using a single word vector. When $n=1$, $v_i$ is the same as word vector $w_i$. Also, the average of fewer words/tokens can be used if the index of word sequences is out of the boundary of an input sentence.}

{\bf{Actions ($a_t$):}} The DAT model's action $a_t$ at time step $t$ is defined as in Figure \ref{fig:Action}. The action gives a transition signal such that the state will change from its current label $l_i$ to its next label $l_j$ by keep the same n-gram vector $v_i$. Simply speaking, the action set $A$ contains all possible labels/tags for the word vector $w_i$ or its correspond n-gram substitute $v_i$, \emph{i.e.} $A=\lbrace l_1, \cdots, l_k\rbrace$. At each time step $t$, the action $a_t$ with highest predicted action-reward value is chosen as:
\begin{equation}
a_t=\argmax_{a} Q(s_t,a|a \in A, \theta_t)
\end{equation}
Then the state will transit from its current label to the label directed by action $a_t$.

{\bf{Rewards ($r_t$):}} The reward defined at a state $s_t$ containing an n-gram vector $v_i$ (with a center word/token $w_i$) will use the one-hot representation $o_{l^*_i}$ of $w_i$'s true label $l^*_i$,  the one hot vector $o_{l_i}$ of the label $l_i$ in current state $s_t$, and the predicted probability $p_i$ for the word $w_i$ using $f_{dnn}$ as:

\begin{equation}
r_t=tanh(log(\dfrac{||o_{l^*_i}-p_i||_2}{||o_{l_i}-o_{l^*_i}||_2+\epsilon}))
\label{reward}
\end{equation}
where $tanh(\cdot)$ is used to normalize the reward to be within -1 and 1, $||\cdot||_2$ is the $\mathcal{L}_2$ norm, and $\epsilon$ is a very small value added to avoid zero denominator.

The insight behind our reward design is using the ratio of the distance from $f_{dnn}$ predicted label to $w_i's$ true label and that from current state's label to its true label. Based on our definition, a higher reward will be assigned to a state in which its label is more closer to the true label compared with the $f_{dnn}$'s predicted one. The reward function is one of the key factors for further improvement using our new augmented tagger. 

{\it{Remarks: It is worth noticing that $p_i$ should be generated by $f_{dnn}$ using the same word sequence as preparing $v_i$. Similarly, $l^*_i$ is the true label of $w_i$ in the same sentence of preparing $p_i$ and $v_i$.}}

{\bf{Training:}}
The training algorithm for DAT is as shown in Algorithm 1. The loss function is defined based on the difference of two estimated expected rewards at state $s_t$ using an unsupervised approach as:
\begin{equation}
\begin{split}
\mathcal{L}_{s_t}&= (\hat{Q}(s_t,a_t|\theta_t)-Q(s_t,a_t|\theta_t))^2\\
&=(r_t+\gamma \max_{a_{t+1}}Q(s_{t+1},a_{t+1}|\theta_t)-Q(s_t,a_t|\theta_t))^2
\end{split}
\label{lossfcn1}
\end{equation}
where $Q(s_t,a_t|\theta_t)$ is the predicted expected reward using $(s_t,a_t)$ generated by the neural network-based estimator in DAT directly, and $\hat{Q}(s_t,a_t|\theta_t)$ is an estimation of $Q(s_t,a_t|\theta_t)$ using the current state reward $r_t$ and its neighbors' estimations $Q(s_{t+1},a_{t+1}|\theta_t)$ as in (\ref{lossfcn1}).

One training technique we borrowed from \cite{mnih2013playing} is the experience replay used in DQN. It improves the convergence issue in neural network-estimator based $Q$-learning by storing the state $s_t$ visited before, action performed $a_t$, state's reward $r_t$ and the next state $s_{t+1}$ after performing the action $a_t$ in an experience tuple ($s_t$,$r_t$, $a_t$, $s_{t+1}$). This tuple is then pushed into the experience replay memory queue $M$. The size of $M$ is pre-defined based on experiment.

Whenever an action is performed then a new state is arrived, the past experience tuple will be pushed into the replay memory queue $M$ if $M$ is not full, otherwise $M$ will pop out the first tuple and push in the latest one as First-in First-out (FIFO). At each training iteration, a random tuple is selected from $M$ and the loss function value is calculated based on the $s_t$, $r_t$, $a_t$ and $s_{t+1}$ given by the tuple. 

\def\algbackskip{\hskip\dimexpr+\labelsep}
\def\LState{\STATE \algbackskip}%

\begin{algorithm}[tb]
\caption{DAT Training using Experience Replay}\label{alg:RL}
\begin{algorithmic}[1]
\STATE Epochs: $N $ 
\STATE Batch Size: $K$ 
\STATE Initialize a replay memory $M$ with size $\mu$ 
\FOR {n=1 $\rightarrow$ N}
\STATE Time step in state space: $t\gets0$
\STATE Randomly select an initial state $s_t=(v_i,l_i)$, $l_i^* \gets$ true label of $v_i$
\WHILE{$l_i$!=$l_i^*$}
\STATE $Q(s_{t},a_{t}\vert\theta_t) \gets $ DAT($s_t$) 
\STATE $a_t$=$\argmax_{a}Q(s_{t},a\vert\theta_t)$
\STATE Perform action $a_t$, generating next state $s_{t+1}$
\STATE 	Push tuple $(s_t,r_t,a_t,s_{t+1})$ into replay memory $M$
\FOR {b=1 $\rightarrow$ $K$}
\STATE Randomly select a tuple $m$ from $M$
\STATE $s_t \gets m(0)$, $r_t \gets m(1)$, $s_{t+1} \gets m(3)$
\STATE $Q(s_{t},a_{t}\vert\theta_t) \gets $ DAT($s_t$)
\STATE $Q(s_{t+1},a_{t+1}\vert\theta_t) \gets$ DAT($s_{t+1}$)
\STATE Obtain $p_i$ from $f_{dnn}$ and $o_{l_i}$ in state $s_t$, $o_{l^*_i}$ from the ground truth in training data 
\STATE $r_t$ $\gets$ $tanh(log(\dfrac{||o_{l^*_i}-p_i||_2}{||o_{l_i}-o_{l^*_i}||_2+\epsilon}))$
\STATE Update $\hat{Q}(s_t,a_t\vert\theta_t)$ in (\ref{lossfcn1})
\STATE Update the network by minimizing loss $\mathcal{L}_{s_t}$ in (\ref{lossfcn1})
\ENDFOR
\STATE $t\gets t+1$
\ENDWHILE\label{euclidendwhile}
\ENDFOR
\end{algorithmic}
\end{algorithm}

{\it{Remarks:}
sIt is worth noticing that it is necessary to use reinforcement learning (RL) in our problem, since RL based model can learn the long-term dependency much better than the ``conventional'' deep learning approach. In our case, since we split a sentence into multiple word level vectors and select them randomly during training, not only we need to consider what the state’s current label is, but also how this state/word level vector is affected by its connected state in the network, such that it won’t have any difficulty to find the correct label from its current wrong label.
\subsubsection{Model Inference}
As shown in Figure \ref{fig:DATInference}, the inference part of the DAT model is a bit different from training the DAT. Since during the training procedure, the entire training dataset ($x_{train}$, $y_{train}$) is used to train both DNN and DAT models. Comparatively, during inference, only partial of test data with ``unsatisfied performance" will be collected and further evaluated by DAT. This is mainly because that we use DAT as an augmented tagger to compensate the minority cases when the original DNN based tagger $f_{dnn}$ does not perform well. For the majority test data $x_{predict\_dnn}$, on which $f_{dnn}$ can perform very well, we still use $f_{dnn}$ to generate their tags $y_{predict\_dnn}$ . In order to filter those data with ``unsatisfied performance", a threshold value $T_r$ is defined. All the tokens with their predicted tags' probabilities below $T_r$ are filtered as the minority cases and further used as the inference input of DAT, \emph{i.e.} $x_{dat}=x_{test \setminus predict\_dnn}$. The outputs of DAT are the actions that will transfer the states from their current labels $l_i$ to the target label $l^*_i$, which gives the output of minority cases, \emph{i.e.} $y_{predict\_dat}$.

{\it{Remarks:} The choice of $T_r$ will slightly affect the performance of DAT as different percentage of data will be filtered. A general recommended $T_r$ is to use a similar percentage of data under minority tags in training data as shown in section \ref{sec-NER}. For example, we use a $T_r$ to filter around 14$\%$(=6,323/45,030) percent of test data in ATIS for DAT since the training tokens under minority tags in ATIS are roughly around this ratio, as shown in Table \ref{table:data_comparison2}.}

\section{Experiment}

\subsection{Data Sets}
In our experiment, we will evaluate our deep reinforcement learning based augmented tagging system on two sequence labeling tasks: Slot filling task and NER task.\\
{\bf{Slot Filling:}} For Slot filling task, we use the public ATIS dataset which  follows the same format as in \cite{liu2015recurrent,mesnil2015using,xu2013convolutional,liu2016attention}. The training set contains 4,978 utterances/sequences, and the test dataset contains 893 utterances. There are totally 127 slot tags.\\
{\bf{NER:}} For NER task, we use the public CoNLL-2003 dataset as in \cite{lample2016neural,chiu2016named,ma2016end,xu2017local,huang2015bidirectional}. The training set contains 14,987 sentences, and the test dataset contains 3,684 sentences. The dataset contains four different types of named entities: \emph{PERSON, LOCATION, ORGANIZATION, and MISC}. By using the BIO2 tagging scheme, the total number of different labels is 9.  There are a total of 204,567 tokens in training set, and 46,666 tokens in test set.
\subsection{Training Setup}

{\bf{Slot Filling:}} For slot filling task, the pre-trained DNN model $f_{dnn}$ has the same set-up as in \cite{liu2016attention}, by using an attention based bi-directional LSTM. The number of states in LSTM cell is 128. The randomly initialized word embedding size is also 128.The batch size is 16 and the dropout rate for non-recurrent connection is 0.5. \\
{\bf{NER:}} For NER task, the pre-trained DNN model $f_{dnn}$ follows the BLSTM-CNNS-CRF structure as in \cite{ma2016end}, which gives the current state-of-the-art result on CoNLL-2003 dataset. The word embedding is chosen as the GloVe 100-dimensional embedding \cite{pennington2014glove}. The CNN's window size is chosen as 3 and the number of filters is 30. The number of states in LSTM cell is 200. The batch size is 10 and the dropout rate is 0.5.

The DAT model $f_{dat}$ used for Slot fill and NER tasks are almost the same except the batch size. The network structure chosen to estimate the action-value function $Q$ is an LSTM structure with 100 states. The averaged word vector in a reinforcement learning state is chosen as a trigram, \emph{i.e.} n=3. The discount factor $\gamma$ in (\ref{lossfcn1}) is selected as 0.5, 0.7 and 0.9 for difference experiments. The minibatch size is $K=16$ for slot filling task and $K=10$ for NER in order to keep the same training batch size as $f_{dnn}$ in different tasks, and the replay memory size is pre-defined as $\mu=5,000$. The thresholds for both experiment are set as $T_r=0.95$.

\subsection{Performance on ATIS dataset}
Our first experiment is performed on ATIS dataset and compared with other existing approaches, by evaluating their slot filling $F1$ scores. A detailed comparison is given in Table \ref{table:comparisonAtis}.

\begin{table}[ht]\scriptsize

\parbox{.45\linewidth}{
\centering
	\caption{Performance of Different Models on ATIS Dataset}
	\label{table:comparisonAtis}
	\begin{tabular}{>{\centering\arraybackslash}p{5.4cm}|>{\centering\arraybackslash}p{1.5cm}}
		\toprule
		
		\multirow{1}{*}{\textbf{Model}} & \multirow{1}{*}{\makecell{\textbf{F1 Score}}}\\
		\midrule
		\midrule
		\multirow{2}{*} {}Recursive NN \cite{guo2014joint} & 93.96\% \\ 
		
		\multirow{2}{*}{}RNN with Label Sampling \cite{liu2015recurrent} & 94.89\%  \\

		\multirow{2}{*}{}Hybrid RNN \cite{mesnil2015using} & 95.06\%   \\		
		\multirow{2}{*}{}RNN-EM \cite{peng2015recurrent} & 95.25\% 	\\
		\multirow{2}{*}{}CNN CRF \cite{xu2013convolutional} & 95.35\%  	\\
		
		\multirow{2}{*}{}Encoder-labeler Deep LSTM \cite{kurata2016leveraging}& 95.66\% 	\\
		\multirow{2}{*}{}Attention Encoder-Decoder NN \cite{liu2016attention}& 95.87\%	\\
		\multirow{2}{*}{}Attention BiRNN \cite{liu2016attention}& 95.98\% 	\\
		\midrule
		\multirow{2}{*}{}DRL based Augmented Tagging System ($\gamma=0.5$) & \textbf{96.85\%} \\
		
		\multirow{2}{*}{}DRL based Augmented Tagging System ($\gamma=0.7$) & \textbf{97.23\%} \\
		\multirow{2}{*}{}DRL based Augmented Tagging System ($\gamma=0.9$) & \textbf{97.86\%}\\

		\bottomrule		
	\end{tabular}
}
\hfill
\parbox{.45\linewidth}{
\centering
	\caption{Performance of Different Models on CoNLL-2003 Dataset}
	\label{table:Comparison_CoNLL2003}
	\begin{tabular}{>{\centering\arraybackslash}p{5.4cm}|>{\centering\arraybackslash}p{1.5cm}}
		\toprule
		
		\multirow{1}{*}{\textbf{Model}} & \multirow{1}{*}{\makecell{\textbf{F1 Score}}}\\
		\midrule
		\midrule
		\multirow{2}{*} {}NN+SLL+LM2\cite{collobert2011natural} & 88.67\% \\ 
		
		\multirow{2}{*}{}NN+SLL+LM2+Gazetter$^*$ \cite{collobert2011natural} & 89.59\%  \\

		\multirow{2}{*}{}BI-LSTM-CRF$^*$ \cite{huang2015bidirectional} & 90.10\%   \\		
		\multirow{2}{*}{}ID-CNN-CRF \cite{strubell2017fast} & 90.54\% 	\\
		\multirow{2}{*}{}FOFE \cite{xu2017local} & 90.71\%  	\\
		\multirow{2}{*}{}BLSTM-CNN+emb \cite{chiu2016named}& 90.91\%	\\
		\multirow{2}{*}{}BLSTM-CRF\cite{lample2016neural}& 90.94\% 	\\
		
		\multirow{2}{*}{}BLSTM-CNN-CRFs \cite{ma2016end}& 91.21\% 	\\
		\multirow{2}{*}{}BLSTM-CNN+emb+lex$^*$ \cite{chiu2016named}& 91.62\% 	\\
		\midrule
		\multirow{2}{*}{}DRL based Augmented Tagging System ($\gamma=0.5$) & \textbf{91.92\%} \\
		
		\multirow{2}{*}{}DRL based Augmented Tagging System ($\gamma=0.7$) & \textbf{92.23\%} \\
		\multirow{2}{*}{}DRL based Augmented Tagging System ($\gamma=0.9$) & \textbf{92.67\%}\\

		\bottomrule		
	\end{tabular}
}
\end{table}
By using the same DNN based model for $f_{dnn}$ as in \cite{liu2016attention}, our new model surpassed the previous state-of-the-art result by 0.9$\%$,1.2$\%$ and  1.9$\%$ for $\gamma$= 0.5, 0.7 and 0.9 separately. It is also worth noticing that, a larger discount factor $\gamma$ gives a better performance as shown in Table. One empirical explanation is that when a larger discount factor is used, the model puts more weights on future states during training, hence it can search the correct label for the current word within a state in a faster manner.
\subsection{Performance on CoNLL-2003 dataset}
Our second experiment is conducted on the CoNLL-2003 dataset and compared with current existing neural network-based approaches for the NER tasks. The metric is also using their F1 scores, and the result is as shown in Table \ref{table:Comparison_CoNLL2003} (The results are collected before initial submission, some better results are released during final submission \cite{peters2018deep}, despite our model's performance is still better).

The methods marked using $*$ are using extra features like lexicons and etc. It can be observed that our DAT system's result outperforms the state-of-the-art results in \cite{ma2016end} (without lexicon features) and \cite{chiu2016named} (lexicon features) by 1$\%$ and 1.4$\%$  using $\gamma=0.9$ separately. Similar to the SLU task, a larger discount factor $\gamma$ also boosts our system's performance.

\subsection{Change on Result Distribution}
Another observation from our experiment's result is that the tags' distribution of wrongly labeled data changed. Table \ref{table:data_comparison3} gives a summary about these changes on ATIS and CoNLL-2003 two test dataset.

\begin{table*}[t]\scriptsize
\centering
	\caption{Change on Tags' Distribution of Wrongly Labeled Data on ATIS and CoNLL-2003 Test Datasets}
	\label{table:data_comparison3}
	\begin{tabular}{>{\centering\arraybackslash}p{3cm}|>{\centering\arraybackslash}p{1cm}>{\centering\arraybackslash}p{1cm}|>{\centering\arraybackslash}p{1cm}>{\centering\arraybackslash}p{1cm}|>{\centering\arraybackslash}p{1cm}>{\centering\arraybackslash}p{1cm}|>{\centering\arraybackslash}p{1cm}>{\centering\arraybackslash}p{1cm}}
		\toprule
		
		\multicolumn{1}{c}{} & \multicolumn{4}{c|}{ATIS Test Dataset} & \multicolumn{4}{c}{\makecell{CoNLL-2003 Test Dataset}}\\
		
		\midrule
		\multirow{1}{*} {}  & \multicolumn{2}{c|}{$\%$ of wrongly labeled data}&\multicolumn{2}{c|}{$\%$ of wrongly labeled data}& \multicolumn{2}{c|}{\makecell{$\%$ of wrongly labeled data}}& \multicolumn{2}{c}{\makecell{$\%$ of wrongly labeled data}}\\
		
		\multirow{1}{*} {}  & \multicolumn{2}{c|}{using $f_{dnn}$}&\multicolumn{2}{c|}{using $f_{dnn}$+$f_{dat}$}& \multicolumn{2}{c|}{\makecell{using $f_{dnn}$}}& \multicolumn{2}{c}{\makecell{using $f_{dnn}$+$f_{dat}$}}\\
		
		\midrule
		\multirow{2}{*} {}Minority Tags  & \multicolumn{2}{c|}{\makecell{92$\%$}}&\multicolumn{2}{c|}{\makecell{56$\%$}}&\multicolumn{2}{c|}{\makecell{78$\%$}}&\multicolumn{2}{c}{\makecell{48$\%$}}\\
		($<$1$\%$ of total $\#$ of data)\\
		\multirow{2}{*}{}Majority Tags  & \multicolumn{2}{c|}{\makecell{8$\%$}}&\multicolumn{2}{c|}{\makecell{44$\%$}}&\multicolumn{2}{c|}{\makecell{22$\%$}}&\multicolumn{2}{c}{\makecell{52$\%$}} \\
		($\geq$ 1$\%$ of total $\#$ of data)\\

		\midrule
		\multirow{2}{*}{}Total & \multicolumn{2}{c|}{\makecell{100$\%$}}&\multicolumn{2}{c|}{\makecell{100$\%$}}&\multicolumn{2}{c|}{\makecell{100$\%$}}&\multicolumn{2}{c}{\makecell{100$\%$}}\\
		
		\bottomrule		
	\end{tabular}
\end{table*}

It can be observed that the percentage of wrongly labeled data with minority tags decreases for both ATIS and CoNLL-2003 datasets by adding the DAT model $f_{dat}$ to the DNN based model $f_{dnn}$. It indicates that the DAT model $f_{dat}$ can help improve the performance of general sequence tagging model by correctly labeling data with  minority tags, which is also the weakness of a single DNN based model $f_{dnn}$.

\section{Conclusion}
In this paper, a new DRL based augmented general tagging system is designed. The system use two sequence labeling models: one is a ``conventional" DNN based tagger, and the other is a novel DRL based augmented tagger, \emph{i.e.} DAT. By filtering DNN's output and picking out the ``unsatisfied data", DAT can further improve sequence labeling tasks' performance by correctly relabeling the data below the threshold $T_r$, especially for those under minority tags. The experiment results on two sequence labeling tasks, \emph{i.e.} Slot fillings and NER in SLU and NLU, both outperform the current state-of-the-art models. Besides the decent experimental performance obtained, more importantly, the new augmented approach can be generalized to more general sequence labeling models without changing their original model setups. 

\bibliographystyle{acl}
\bibliography{2018Coling}

\begin{thebibliography}{}

\bibitem[\protect\citename{Chawla \bgroup et al.\egroup
  }2004]{chawla2004special}
Nitesh~V Chawla, Nathalie Japkowicz, and Aleksander Kotcz.
\newblock 2004.
\newblock Special issue on learning from imbalanced data sets.
\newblock {\em ACM Sigkdd Explorations Newsletter}, 6(1):1--6.

\bibitem[\protect\citename{Chiu and Nichols}2016]{chiu2016named}
Jason~PC Chiu and Eric Nichols.
\newblock 2016.
\newblock Named entity recognition with bidirectional lstm-cnns.
\newblock {\em Transactions of the Association for Computational Linguistics},
  4:357--370.

\bibitem[\protect\citename{Collobert \bgroup et al.\egroup
  }2011]{collobert2011natural}
Ronan Collobert, Jason Weston, L{\'e}on Bottou, Michael Karlen, Koray
  Kavukcuoglu, and Pavel Kuksa.
\newblock 2011.
\newblock Natural language processing (almost) from scratch.
\newblock {\em Journal of Machine Learning Research}, 12(Aug):2493--2537.

\bibitem[\protect\citename{Guo \bgroup et al.\egroup }2014]{guo2014joint}
Daniel Guo, Gokhan Tur, Wen-tau Yih, and Geoffrey Zweig.
\newblock 2014.
\newblock Joint semantic utterance classification and slot filling with
  recursive neural networks.
\newblock In {\em Spoken Language Technology Workshop (SLT), 2014 IEEE}, pages
  554--559. IEEE.

\bibitem[\protect\citename{He and Garcia}2009]{he2009learning}
Haibo He and Edwardo~A Garcia.
\newblock 2009.
\newblock Learning from imbalanced data.
\newblock {\em IEEE Transactions on knowledge and data engineering},
  21(9):1263--1284.

\bibitem[\protect\citename{He \bgroup et al.\egroup }2008]{he2008adasyn}
Haibo He, Yang Bai, Edwardo~A Garcia, and Shutao Li.
\newblock 2008.
\newblock Adasyn: Adaptive synthetic sampling approach for imbalanced learning.
\newblock In {\em Neural Networks, 2008. IJCNN 2008.(IEEE World Congress on
  Computational Intelligence). IEEE International Joint Conference on}, pages
  1322--1328. IEEE.

\bibitem[\protect\citename{Hemphill \bgroup et al.\egroup
  }1990]{hemphill1990atis}
Charles~T Hemphill, John~J Godfrey, George~R Doddington, et~al.
\newblock 1990.
\newblock The atis spoken language systems pilot corpus.
\newblock In {\em Proceedings of the DARPA speech and natural language
  workshop}, pages 96--101.

\bibitem[\protect\citename{Huang \bgroup et al.\egroup
  }2015]{huang2015bidirectional}
Zhiheng Huang, Wei Xu, and Kai Yu.
\newblock 2015.
\newblock Bidirectional lstm-crf models for sequence tagging.
\newblock {\em arXiv preprint arXiv:1508.01991}.

\bibitem[\protect\citename{Kurata \bgroup et al.\egroup
  }2016]{kurata2016leveraging}
Gakuto Kurata, Bing Xiang, Bowen Zhou, and Mo~Yu.
\newblock 2016.
\newblock Leveraging sentencelevel information with encoder lstm for natural
  language understanding.
\newblock {\em arXiv preprint}.

\bibitem[\protect\citename{Lample \bgroup et al.\egroup
  }2016]{lample2016neural}
Guillaume Lample, Miguel Ballesteros, Sandeep Subramanian, Kazuya Kawakami, and
  Chris Dyer.
\newblock 2016.
\newblock Neural architectures for named entity recognition.
\newblock {\em arXiv preprint arXiv:1603.01360}.

\bibitem[\protect\citename{Liu and Lane}2015]{liu2015recurrent}
Bing Liu and Ian Lane.
\newblock 2015.
\newblock Recurrent neural network structured output prediction for spoken
  language understanding.
\newblock In {\em Proc. NIPS Workshop on Machine Learning for Spoken Language
  Understanding and Interactions}.

\bibitem[\protect\citename{Liu and Lane}2016]{liu2016attention}
Bing Liu and Ian Lane.
\newblock 2016.
\newblock Attention-based recurrent neural network models for joint intent
  detection and slot filling.
\newblock {\em arXiv preprint arXiv:1609.01454}.

\bibitem[\protect\citename{Ma and Hovy}2016]{ma2016end}
Xuezhe Ma and Eduard Hovy.
\newblock 2016.
\newblock End-to-end sequence labeling via bi-directional lstm-cnns-crf.
\newblock {\em arXiv preprint arXiv:1603.01354}.

\bibitem[\protect\citename{Mesnil \bgroup et al.\egroup }2015]{mesnil2015using}
Gr{\'e}goire Mesnil, Yann Dauphin, Kaisheng Yao, Yoshua Bengio, Li~Deng, Dilek
  Hakkani-Tur, Xiaodong He, Larry Heck, Gokhan Tur, Dong Yu, et~al.
\newblock 2015.
\newblock Using recurrent neural networks for slot filling in spoken language
  understanding.
\newblock {\em IEEE/ACM Transactions on Audio, Speech and Language Processing
  (TASLP)}, 23(3):530--539.

\bibitem[\protect\citename{Mnih \bgroup et al.\egroup }2013]{mnih2013playing}
Volodymyr Mnih, Koray Kavukcuoglu, David Silver, Alex Graves, Ioannis
  Antonoglou, Daan Wierstra, and Martin Riedmiller.
\newblock 2013.
\newblock Playing atari with deep reinforcement learning.
\newblock {\em arXiv preprint arXiv:1312.5602}.

\bibitem[\protect\citename{Mnih \bgroup et al.\egroup }2015]{mnih2015human}
Volodymyr Mnih, Koray Kavukcuoglu, David Silver, Andrei~A Rusu, Joel Veness,
  Marc~G Bellemare, Alex Graves, Martin Riedmiller, Andreas~K Fidjeland, Georg
  Ostrovski, et~al.
\newblock 2015.
\newblock Human-level control through deep reinforcement learning.
\newblock {\em Nature}, 518(7540):529.

\bibitem[\protect\citename{Peng and Yao}2015]{peng2015recurrent}
Baolin Peng and Kaisheng Yao.
\newblock 2015.
\newblock Recurrent neural networks with external memory for language
  understanding.
\newblock {\em arXiv preprint arXiv:1506.00195}.

\bibitem[\protect\citename{Pennington \bgroup et al.\egroup
  }2014]{pennington2014glove}
Jeffrey Pennington, Richard Socher, and Christopher Manning.
\newblock 2014.
\newblock Glove: Global vectors for word representation.
\newblock In {\em Proceedings of the 2014 conference on empirical methods in
  natural language processing (EMNLP)}, pages 1532--1543.

\bibitem[\protect\citename{Peters \bgroup et al.\egroup }2018]{peters2018deep}
Matthew Peters, Mark Neumann, Mohit Iyyer, Matt Gardner, Christopher Clark,
  Kenton Lee, and Luke Zettlemoyer.
\newblock 2018.
\newblock Deep contextualized word representations.
\newblock In {\em Proceedings of the 2018 Conference of the North American
  Chapter of the Association for Computational Linguistics: Human Language
  Technologies, Volume 1 (Long Papers)}, volume~1, pages 2227--2237.

\bibitem[\protect\citename{Powell}2007]{powell2007approximate}
Warren~B Powell.
\newblock 2007.
\newblock {\em Approximate Dynamic Programming: Solving the curses of
  dimensionality}, volume 703.
\newblock John Wiley \& Sons.

\bibitem[\protect\citename{Strubell \bgroup et al.\egroup
  }2017]{strubell2017fast}
Emma Strubell, Patrick Verga, David Belanger, and Andrew McCallum.
\newblock 2017.
\newblock Fast and accurate sequence labeling with iterated dilated
  convolutions.
\newblock {\em arXiv preprint arXiv:1702.02098}.

\bibitem[\protect\citename{Sun \bgroup et al.\egroup
  }2009]{sun2009classification}
Yanmin Sun, Andrew~KC Wong, and Mohamed~S Kamel.
\newblock 2009.
\newblock Classification of imbalanced data: A review.
\newblock {\em International Journal of Pattern Recognition and Artificial
  Intelligence}, 23(04):687--719.

\bibitem[\protect\citename{Wang \bgroup et al.\egroup }2018]{wang2018bi}
Yu~Wang, Yilin Shen, and Hongxia Jin.
\newblock 2018.
\newblock A bi-model based rnn semantic frame parsing model for intent
  detection and slot filling.
\newblock In {\em Proceedings of the 2018 Conference of the North American
  Chapter of the Association for Computational Linguistics: Human Language
  Technologies, Volume 2}, volume~2, pages 309--314.

\bibitem[\protect\citename{Xu and Sarikaya}2013]{xu2013convolutional}
Puyang Xu and Ruhi Sarikaya.
\newblock 2013.
\newblock Convolutional neural network based triangular crf for joint intent
  detection and slot filling.
\newblock In {\em Automatic Speech Recognition and Understanding (ASRU), 2013
  IEEE Workshop on}, pages 78--83. IEEE.

\bibitem[\protect\citename{Xu \bgroup et al.\egroup }2017]{xu2017local}
Mingbin Xu, Hui Jiang, and Sedtawut Watcharawittayakul.
\newblock 2017.
\newblock A local detection approach for named entity recognition and mention
  detection.
\newblock In {\em Proceedings of the 55th Annual Meeting of the Association for
  Computational Linguistics (Volume 1: Long Papers)}, volume~1, pages
  1237--1247.

\end{thebibliography}

%
%
%
%
%
%

\end{document}